\documentclass[letterpaper, 10 pt, conference]{ieeeconf}  % Comment this line out if you need a4paper

\IEEEoverridecommandlockouts                              % This command is only needed if 
\title{\LARGE \bf
A Comprehensive Analysis of the Effects of Network Quality of Service on Robotic Telesurgery
}

\author{Zhaomeng Zhang$^{\dagger1}$, Seyed Hamid Reza Roodabeh$^{\dagger1}$, Homa Alemzadeh$^{1}$
\thanks{$\dagger$ These authors contributed equally and are the corresponding authors.}
\thanks{This work was supported in part by the the National Science Foundation (NSF) grants CNS-2146295 and CCF-2402941.}
\thanks{$^{1}$Department of Electrical and Computer Engineering, University of Virginia, Charlottesville, VA 22093, USA {\tt\small \{sus7sv,ydq9ag,ha4d\}@virginia.edu}}%
\thanks{The source code and dataset used for this paper are available in our GitHub repository: \url{https://github.com/UVA-DSA/telesurgery-qos-analysis}.}
}
\usepackage{caption}
\usepackage{graphicx}
\usepackage{algorithm}
\usepackage{algorithmic}
\usepackage{amsmath}
\usepackage{amssymb}
\usepackage{subcaption}
\usepackage{booktabs}
\usepackage{multirow}
\usepackage{todonotes}
\usepackage{comment}
\usepackage{url}
\newcommand{\network}[1]{{{\color{black}#1}}{}}

\usepackage{hyperref}

\makeatletter
\def\bstctlcite{\@ifnextchar[{\@bstctlcite}{\@bstctlcite[@auxout]}}
\def\@bstctlcite[#1]#2{\@bsphack
  \immediate\write\csname #1\endcsname{\string\citation{#2}}\@esphack}
\makeatother

\begin{document}

\bstctlcite{IEEEexample:BSTcontrol}

\maketitle
\thispagestyle{empty}
\pagestyle{empty}

%%%%%%%%%%%%%%%%%%%%%%%%%%%%%%%%%%%%%%%%%%%%%%%%%%%%%%%%%%%%%%%%%%%%%%%%%%%%%%%%
\begin{abstract}

The viability of long-distance telesurgery hinges on reliable network Quality of Service (QoS), yet the impact of realistic network degradations on task performance is not sufficiently understood. This paper presents a comprehensive analysis of how packet loss, delay, and communication loss affect telesurgical task execution. We introduce NetFI, a novel fault injection tool that emulates different network conditions using stochastic QoS models informed by real-world network data. By integrating NetFI with a surgical simulation platform, we conduct a user study involving 15 participants at three proficiency levels, performing a standardized Peg Transfer task under varying levels of packet loss, delay, and communication loss. We analyze the effect of network QoS on overall task performance and the fine-grained motion primitives (MPs) using objective performance and safety metrics and subjective operator's perception of workload. We identify specific MPs vulnerable to network degradation and find strong correlations between proficiency, objective performance, and subjective workload. These findings offer quantitative insights into the operational boundaries of telesurgery. Our open-source tools and annotated dataset provide a foundation for developing robust and network-aware control and mitigation strategies.

\end{abstract}

%%%%%%%%%%%%%%%%%%%%%%%%%%%%%%%%%%%%%%%%%%%%%%%%%%%%%%%%%%%%%%%%%%%%%%%%%%%%%%%%
\section{INTRODUCTION}

Robotic Assisted Minimally Invasive Surgery (RAMIS) has seen widespread clinical adoption, exemplified by the da Vinci surgical system, which has been used in over 17 million procedures worldwide\cite{intuitive_report_2024}. These systems enhance surgeon dexterity and visual feedback, thereby improving patient outcomes, accelerating recovery, and reducing complications. Conventionally, the master tool manipulators (MTMs) and patient-side manipulators (PSMs) are co-located in the operating room, linked by a wired connection to ensure robust, low-latency data transmission. However, the emergence of high-bandwidth, low-latency 5G/6G networks is overcoming this limitation, enabling the paradigm of long-distance telesurgery \cite{Vipul}.
Telesurgery could reduce the need for physical presence while increasing the flexibility of surgical resources in extreme or high-risk environments, such as rural areas, battlefields \cite{garcia}, deep sea \cite{Doarn}, and  outer space \cite{Pantalone}. Although several long-distance (up to 1000km) procedures have been successfully conducted in the past \cite{marescaux2002}, challenges related to network communication stability remain a major barrier to the wide adoption of robotic telesurgery \cite{Yip}. 

Network \network{Quality of Service (QoS) degradations} such as delay, packet loss, and communication loss can significantly impact surgeons' intra-operative decisions and performance, potentially leading to unsafe situations and medical complications. Several studies have examined the impact of varying types and degrees of degradations on the feasibility and user performance in surgical teleoperation \cite{Wang2025}. These studies indicate that while some surgeons consider network latencies below 200ms to be acceptable \cite{nankaku2022}, latencies as low as 50ms \cite{Yip} to 135ms \cite{kumcu} can still hinder telesurgery performance. Moreover, a 2000km teleoperation study showed that packet losses of 3\% to 7\% under 150 Mbps communication bandwidth can result in corrupted image frames, distracting operators and affecting their performance \cite{ebihara}. 

Furthermore, simulated robots and environments are used in place of physical hardware to emulate QoS degradations, mitigating risks to expensive surgical robots and equipment while enabling flexible task and scenario design~\cite{xu, ishida, ouda2025}. QoS degradation has also been studied in digital twin architectures \cite{ishida, gonzalez, bonne, wang} and cloud robotics \cite{chen2024cloud} to support intent recognition, autonomy, and fault-tolerance for mitigating adverse events. 

%Earlier works also studied the effect of simulated network-based attacks that intercept and compromise packets in order to modify or manipulate the operator's intent or hijack the robot~\cite{Bonaci2015}, which can potentially have effects similar to severe QoS degradations. 

%Given that network degradations can pose safety risks and hazards to patients in robotic surgery, some research has framed diverse QoS degradations as security threats from man-in-the-middle attacks \cite{Bonaci2015}.

To our knowledge, no prior work has investigated QoS degradation effects using both objective and subjective performance and safety metrics with realistic, model-based emulation of real-world network conditions across severity levels, from near-ideal to intermediate and severe scenarios, which arise in settings like battlefields or remote rural areas. Moreover, no prior work has quantified these effects at the subtask level. Decomposing complex operations into fundamental motion primitives (MPs) \cite{nwoye2020, Hutchinson_2023} provides a generalizable framework for more granular analysis, such as examining the effects of degradations on intricate tool-tissue interactions, development of targeted mitigation strategies, and fine-grained skill assessment. %~\cite{hutchinson2023towards}.

%%%%%%%%%%%%%%%%%%%%
% The assistance of telesurgical robotics simulation offers one of important ways to eliminates the risk of damaging costly surgical robots during the experimentation and also can mitigate the effects of network QoS degradations. For example, in order to determine the acceptable latency levels in telesurgery, previous work \cite{xu} investigates on the different latency effects on the surgical performance using the dV-Trainer (Mimic Technologies, Inc., Seattle, WA) simulator. Furthermore, simulation-based digital twin methods \cite{gonzalez, bonne, ishida, wang} have been proposed to enhance the telesurgical performance under poor QoS conditions, which relies on the local simulated surgical robotic environment to reserve user’s operation information during network outage and send high-level or augmented commands to real robots when communication is recovered. 

% However, two major gaps still persists in the current research of network QoS in telesurgerical simulation. First, there is a lack of comprehensive and realistic network QoS emulators used for telesurgical systems. Second, the impact of network QoS degradations on human motion primitives (MPs) level \cite{Hutchinson_2022} \cite{Hutchinson_2023} in telesurgery has not been systematically investigated from prior research. Filling these gaps will pave the way for the development and validation of more robust solutions to address network challenges in long-distance telesurgery. 

\begin{figure*}
    \centering
    \includegraphics[width=0.86 \linewidth]{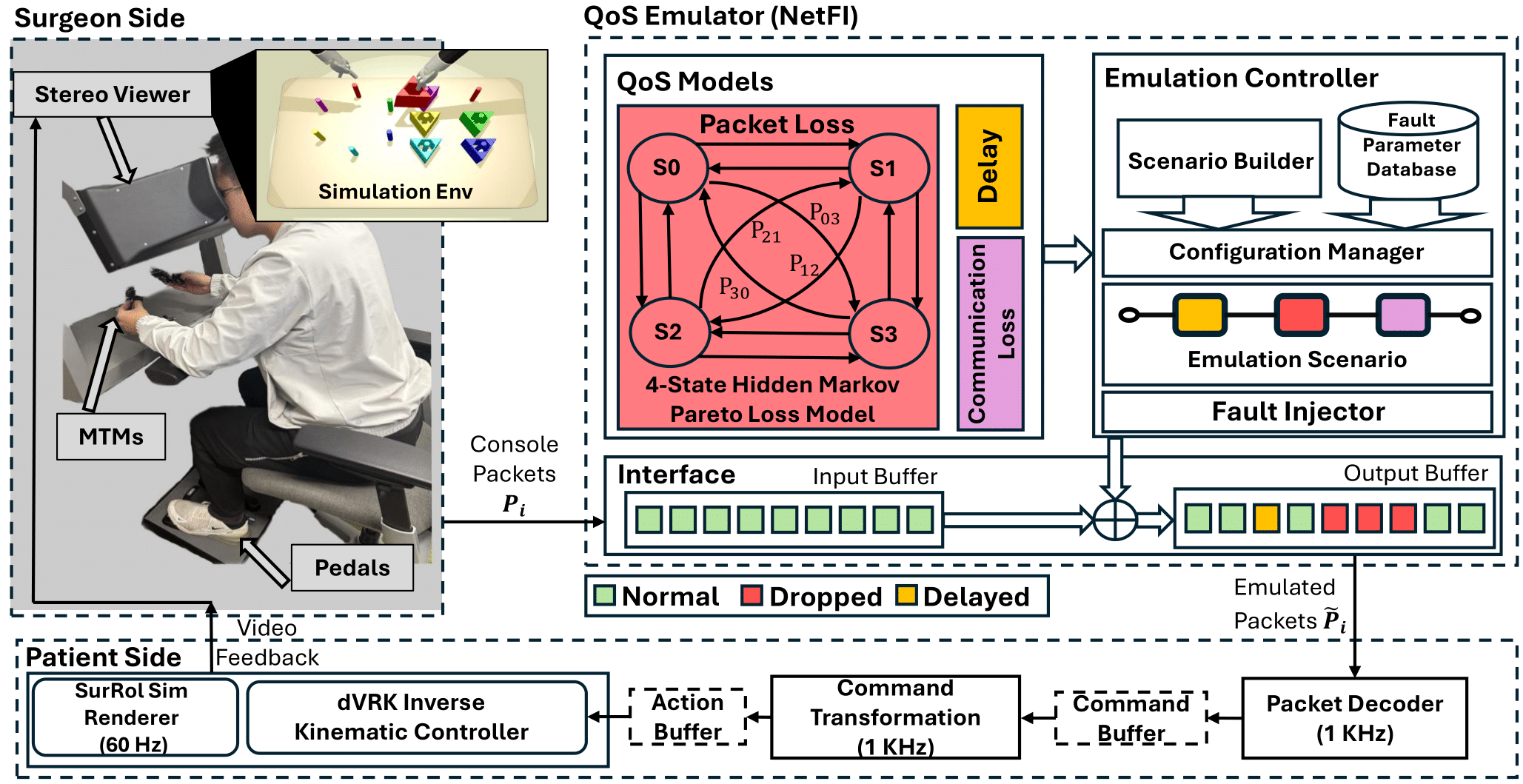}
    \caption{System Architecture of our telesurgical simulation setup with model-based network QoS emulation: the block diagram illustrates the data flow between the surgeon side console, the patient side robot simulator, and the network QoS emulator.}
    \label{fig:system overview}
    \vspace{-1.5em}
\end{figure*}

To address these gaps, we perform a comprehensive analysis of the effects of QoS degradations on task performance, safety, and user experience. Our contributions are as follows: 

\begin{itemize}
\item Introducing a novel model-based \textbf{Network Fault Injection tool (NetFI)} that emulates the effect of communication loss, packet loss, and delay based on realistic models and data from 4G/5G networks, that can be easily integrated into any teleoperation system to simulate diverse network QoS conditions by modifying the stochastic models and their parameters. 

\item Conducting a \textbf{comprehensive user study} involving 15 participants at three proficiency levels, performing a standard Fundamentals of Laparoscopic Surgery (FLS)~\cite{FLS} Peg Transfer task under different network conditions, using an open-source telesurgical simulation system which integrates a state-of-the-art surgical robot simulator (SurRoL~\cite{xusurrol}) with a surgeon console (dVTrainer, Mimic Technologies) and mechanisms for real-time logging of kinematic, video, and foot pedal data for performance evaluation. This study resulted in a \textbf{multi-modal dataset with MP and error labels} of 180 Peg Transfer trials.

    % \item Providing \textbf{new insights} into the effects of different QoS degradation scenarios on \textbf{user performance and operation safety} at both the \textbf{task and MP levels} and across \textbf{user proficiency levels} and the user experience.

    \item Providing \textbf{new insights} into the effects of different QoS degradation scenarios on \textbf{user performance and operation safety}, at both the \textbf{task and MP levels}, for different \textbf{user proficiency levels} and user experience.

\end{itemize}
%The NetFI tool and the annotated dataset from our study is publicly available at [Anonymous URL]. 

\section{BACKGROUND AND RELATED WORK}
\subsection{Networks in Robotic Telesurgery}  

Safe and reliable long-distance telesurgery demands stringent network performance~\cite{patel2025}, motivating numerous studies on how degradations affect operator performance and experience. Nankaku et al.~\cite{nankaku2022} showed that delays $\leq$100ms generally preserve performance, with experienced surgeons maintaining proficiency even at longer delays. Anvari et al.~\cite{anvari2005} reported that latencies $>$300ms significantly increase task completion times and error rates. Feasibility of teleoperation has also been studied with challenging infrastructures such as satellite links in remote or extreme environments~\cite{satellite1} and battlefields~\cite{battlefield}. 

Simulation-based works have further complemented these findings. Xu et al.~\cite{xu} examined the effects of latency on surgical performance using the dV-Trainer. Ouda et al.~\cite{ouda2025} demonstrated that packet loss substantially increases task completion time and error rates, and degrades video quality. Ishida et al.~\cite{ishida} analyzed user responses to randomized communication losses of 0-3s using the AMBF simulator~\cite{munawar1}, but their study did not investigate the impact of comprehensive network degradation on granular action-level performance in dual-arm FLS tasks.

We advance the state-of-the-art in this area through a comprehensive evaluation of performance, safety, and user experience across different network degradation scenarios. Our approach integrates realistic 4G/5G-informed QoS models, diverse error and failure types, performance metrics at the subtask level, and a comparative analysis across three user proficiency levels.

% We enhance the state-of-the-art in this area by performing a comprehensive analysis of task performance, safety, and user experience under different network degradation scenarios, by incorporating realistic QoS models informed by 4G/5G network data, diverse error and failure types, performance metrics at the subtask level, and user experience across three proficiency levels. 

%No prior studies have investigated the effects of low QoS networks on user performance, types and rate of errors occurring, and overall user experience using realistic models of impariments in a quantifiable and structured way, a gap we address in this work. Thus, we propose an open source QoS impairment emulator, NetFI, that can emulate degradations of arbitrary strength, from near-ideal to extreme conditions like those encountered in a battlefield zone or space vehicle, using real-world models of degradations. We use this emulator to conduct a comprehensive user study using a high fidelity surgical simulator and custom teleoperation setup.

\subsection{Peg Transfer and Motion Primitives}
The FLS Peg Transfer (PT) evaluates bimanual dexterity, hand–eye coordination, and depth perception. This task involves transferring six pegs across a pegboard using two graspers. Its simplicity and structured design make it a standard benchmark for assessing surgical skill and training progression. A successful trial of PT involves transferring all pegs efficiently via a mid-air handover without dropping them, moving the graspers out of the camera view or causing any collisions with the other instruments or the environment.
 
\begin{figure}[t!]
    \centering
    \includegraphics[width=1 \linewidth]{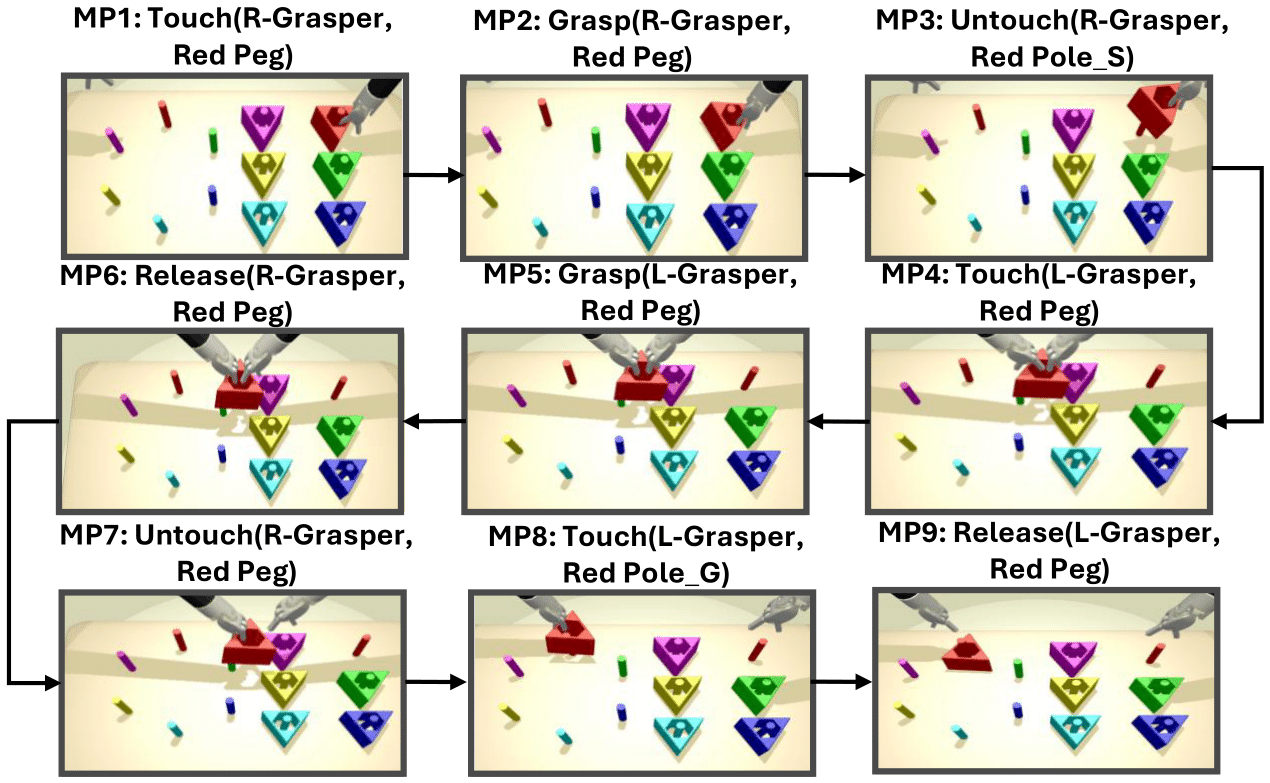}
    \caption{PT Task Workflow: Transfer of a red peg from the start pole to the goal pole in the simulation environment, modeled as a sequence of MPs, MP1 to MP9. ‘L': Left, ‘R’: Right, ‘S’: Start, and ‘G’: Goal.}
    \label{fig:Peg Transfer MPs}
    \vspace{-1em}
\end{figure}

To standardize surgical training and improve the understanding of tool–tissue interactions, prior work~\cite{Neumuth_2011, Hutchinson_2023} has proposed surgical process models that decompose procedures into hierarchical structures, consisting of steps, tasks, gestures, and MPs. Each MP represents a fundamental atomic action (e.g., touch, push, pull, and grasp) involving an instrument and a target object. This framework enables a more fine-grained analysis of user performance and operation safety and has been shown to be generalizable to variety of tasks and complex procedures. %similar to the concept of action triplets~\cite{nwoye2020}, using this framework, user performance and operation safety can be analyzed at a more granular level, which is generalizable to more complex procedures. 
PT, for instance, can be decomposed into nine MPs as shown in Fig. \ref{fig:Peg Transfer MPs}.  
\section{METHODS}

\subsection{Teleoperation Framework}
Our teleoperation setup (Fig. \ref{fig:system overview}) consists of three main components: the surgeon-side console, the NetFI QoS emulator, and the patient-side simulation environment. At the console, a human operator uses MTMs and a clutch pedal to control the simulated PSMs, while viewing the simulated scene through a stereo viewer. The MTMs generate pose change commands for the robotic arms, and the clutch allows hand repositioning for adjustment and re-orientation in the user workspace. These commands are encapsulated into packets $P_i$ using the Interoperable Telesurgical Protocol (ITP)~\cite{ITP} and transmitted via UDP at $\sim$1 kHz. 

The NetFI emulator intercepts and applies a target set of degradations to these packets, producing $\Tilde{P_i}$, which are forwarded to the patient-side robot simulator. We developed a multi-threaded architecture to enable real-time packet processing and rendering on the patient-side simulator, incorporating comprehensive data logging mechanisms for analysis. Our architecture consists of a \texttt{Packet Decoder} thread that decodes and stores incoming packets into a buffer and a high-frequency \texttt{command transformation} thread that accumulates delta pose commands and transforms them into the simulated PSM coordinate system. The \texttt{dVRK Inverse Kinematic Controller} of the SurRoL simulator then retrieves the buffered PSM action commands at a lower rate and converts them into corresponding joint angles, enabling precise execution of the operator’s commands on the simulated PSMs within the virtual 3D environment (see Fig. \ref{fig:system overview}). This design preserves high-frequency user inputs even when the simulation runs at a lower rate in real time. The SurRoL-based simulator~\cite{xusurrol} renders the environment at 60 Hz, with video feedback streamed back to the console via TCP. Unlike the control signals, the video stream is not subjected to NetFI-induced QoS degradations in this study.

\subsection{NetFI QoS Emulator}
The NetFI QoS Emulator is a fault injection tool designed to realistically simulate the effect of different network faults that result in QoS degradation. It uses stochastic models for packet loss, delay, and communication loss that are parameterized to match the characteristics observed in commercial wireless and cellular networks. To facilitate a broad range of research into system safety and security, the emulator is engineered to integrate with any teleoperation setup, allowing degradation scenarios to be studied in isolation or in composition. %for generating task-related faults.

The parameters for each model, which shape characteristics like the mean and skewness of the underlying statistical distributions, are pre-calculated via optimization (see section \ref{optimization}). This process creates a large database of parameters that can reproduce a wide spectrum of real-world network conditions, from ideal to worst-case scenarios (e.g., delay from 0-1300ms, packet loss from 0-95\%). These parameters are stored in a \textit{Fault Parameter Database}. As shown in Fig.~\ref{fig:system overview}, when a user defines an injection campaign (e.g., 50ms delay with 50\% packet loss) via a specific format acceptable by the \textit{Scenario Builder}, the \texttt{Configuration Manager} loads the appropriate stochastic models (see section \ref{models}) and pre-calculated parameters to construct the emulation scenario. This scenario may consist of one or more than one degradation type, which is further structured as an emulation pipeline. Finally, the \texttt{Fault Injector} intercepts packets, which are temporarily held in an input buffer. The active emulation model assigns a degradation status (e.g., normal, dropped, and/or delayed) to each packet. Packets are then placed in an output buffer and relayed to the patient side simulator according to this status. The emulator is implemented using efficient low-level software, ensuring that its own processing overhead adds less than 1 ms of delay.

\subsection{Models of Network Degradation}
\label{models}
To realistically emulate network conditions, we use the following stochastic models for packet loss, delay, and communication loss, which capture the variable and uncertain nature of wireless and cellular networks.

\subsubsection{Packet Loss}
Severe packet loss in wireless networks is often bursty rather than fixed-rate and independently distributed, a phenomenon that can significantly degrade performance in real-time applications like telesurgery\cite{ebihara, ouda2025}. This is true for hospital WLANs, where interference causes burst errors, and for 4G/5G cellular networks under adverse conditions like deep fading, congestion, or handover, where burst losses can last for hundreds of milliseconds \cite{kamtam2024network}. To capture these dynamics, we employ a four-state Gilbert-Elliott (GE) Markov model \cite{ge_og}, which has been validated for both WLAN and LTE networks \cite{packetloss_ge_lte}. The model consists of a `Good' state ($S_0$) with minimal loss, a `Bad' state ($S_1$) with significant loss, and two intermediate states ($S_2$, $S_3$) (see Fig. \ref{fig:system overview}). The probability of transitioning from state $S_i$ to state $S_j$ at any given time is denoted by $P_{ij}$ and the probabilities of transitioning from any given state $S_i$ to all possible next states sum to one, for all $i \in \{0, 1, 2, 3\}$.

To model the heavy-tailed nature of burst loss lengths (BLL) in each state $S_i$, we use a Pareto Type II (Lomax) distribution \cite{packetloss_pareto} with the probability density function: %of the BLL, $x$, is:
\begin{equation}
f(x) = \frac{\alpha_i \lambda_i^{\alpha_i}}{(x + \lambda_i)^{\alpha_i+1}}, \quad x > 0, \quad \alpha_i, \lambda_i > 0
\vspace{-2mm}
\end{equation}

where $\alpha_i$ and $\lambda_i$ are the shape and scale parameters. BLL ($x$) samples are generated using inverse transform sampling:
\begin{equation}
X = \lambda_i \left( (1 - U)^{-\frac{1}{\alpha_i}} - 1 \right)
\vspace{-2mm}
\end{equation}
where $U \sim U(0,1)$ is from a uniform distribution.

To achieve a target loss rate, the NetFI's configuration manager loads the appropriate values of $\alpha_i$, $\lambda_i$ and the transition probabilities $P_{ij}$ for all states $i$. We only optimize the Lomax parameters for good and bad states (see Section \ref{optimization}), and the rest are adopted from~\cite{packetloss_pareto, ge_stat}.

\subsubsection{Delay}
Communication Delay in 4G/5G networks can be highly variable and can be accurately modeled using hyperexponential distributions, which capture scenarios with multiple cascaded delay processes arising from factors like network congestion \cite{hyper-hypo, salah2017performance}. We model the total delay $D$ as $D_{\text{min}} + X$, which incorporates a minimum baseline delay $D_{\text{min}}$ to account for hardware and transmission overhead, plus a random variable $X$ with a hyperexponential distribution. %Our model incorporates a minimum baseline delay ($D_{\text{min}}$) to account for hardware and transmission overhead, plus a random variable $X$ drawn from a hyperexponential distribution. The total delay $D$ is:
%\begin{equation}
%D = D_{\text{min}} + X
%\end{equation}
The probability density function (PDF) for X is a weighted sum of $n$ exponential distributions:
\vspace{-1mm}
\begin{equation}
f(x) = \sum_{i=1}^{n} w_i \lambda_i e^{-\lambda_i x}, \quad x \geq 0
\end{equation}
where $w_i$ are the weights ($\sum w_i = 1$) and $\lambda_i$ are the rate parameters. These parameters are loaded by the configuration manager to achieve a target mean delay, given by:
\begin{equation}
E[D] = D_{\text{min}} + \sum_{i=1}^{n} \frac{w_i}{\lambda_i}
\vspace{-1mm}
\end{equation}

In software, NetFI delays MTM command packets by holding them in a time-scheduled buffer before release. The patient-side controller then executes these time-shifted commands in arrival order, creating a temporal mismatch with the current MTM input.

\subsubsection{Communication Loss}
Extended communication loss is a worst-case scenario in which the teleoperated robot does not receive new packets, representing a total loss of communication. Telesurgical systems typically engage a fail-safe mode or emergency stop to halt the robot operation. Our model simulates these events with three components:
\begin{itemize}
    \item \textbf{Loss Initiation}: Each packet has a probability $p_{\text{loss}}$ of triggering a loss period.
    \item \textbf{Loss Duration}: The duration of the loss, $L$, is sampled from a uniform distribution, $L \sim U(L_{\min}, L_{\max})$. All packets are dropped during this interval.
    \item \textbf{Cooldown Period}: Following a loss event, a cooldown period of $C$ packets is enforced during which no new loss events are initiated, preventing unrealistic consecutive losses.
\end{itemize}
The expected loss rate is the ratio of the average loss duration to the total cycle time (loss + cooldown):
\begin{equation}
\text{Loss Rate} = \frac{E[L]}{E[L] + C} = \frac{(L_{\min} + L_{\max})/2}{(L_{\min} + L_{\max})/2 + C}
\end{equation}

Our configuration manager selects the values of $L_{min}$, $L_{max}$ and $C$ to achieve a target loss rate.

\subsection{Emulator Parameter Optimization}
\label{optimization}
We use Monte-Carlo simulation~\cite{MC_OPT} and differential evolution optimization~\cite{differential_evolution} to find the optimal values for the tunable parameters in each degradation model in order to meet target metrics (specific packet or communication loss rate, delay). We first gather QoS measurements and performance benchmarks for target networks (e.g., 4G, 5G, Wi-Fi, Satellite) from commercial network studies \cite{QoS_Stats1, QoS_Stats2, QoS_Stats3} to define realistic QoS degradation severity levels for routine and critical scenarios.

Then for each model with the parameter vector $\theta = [\theta_1, \theta_2, \dots, \theta_n]$ (e.g. $\theta_{delay}=[w_1, \lambda_1, ..., w_n, \lambda_n, n]$), we minimize the squared error between a target degradation level (e.g., 100ms delay), $M_{target}$, and the empirical metric obtained from a Monte-Carlo simulation $M_{\text{emp}}(\theta)$:
\[
J(\theta) = \left( M_{\text{emp}}(\theta) - M_{\text{target}} \right)^2
\]

This multivariate optimization problem is solved numerically using differential evolution and this process is repeated to generate a \textit{parameter database} for our study.  
%to find the parameter vector $\theta$ that best reproduces the desired network behavior. This process is repeated to generate parameter sets for different impairment severities.

\section{EXPERIMENTAL SETUP}

To investigate the impact of network QoS degradations on telesurgical performance, we designed a user study to answer the following research questions:

\noindent \textbf{RQ1:} How do different types and severities of network degradations affect overall task performance and safety?

\noindent \textbf{RQ2:} How do these degradations affect the execution of individual MPs?

\noindent \textbf{RQ3:} How do different network degradations impact the subjective user experience and perceived workload?

\subsection{User Study Design}
We conducted a user study with 15 graduate engineering students (12 male, 3 female; age: $26.7\pm4.4$), approved by the university's Institutional Review Board (IRB). Initially, participants underwent a familiarization phase to practice the control of the surgeon console. They proceeded to the main experiment only after demonstrating proficiency by successfully completing the PT task under normal network conditions. Following familiarization, each participant completed a System Usability Scale (SUS) \cite{brooke1996} questionnaire to evaluate the baseline usability of our system.
% to establish a baseline usability score. 
This ensured that performance results reflected network conditions and skill rather than interface-related difficulties.

%Based on their average single-peg transfer time in the final trial under normal conditions, participants were classified into three proficiency levels: \textbf{Novice} ($t > 25\,\mathrm{s}$), \textbf{Intermediate} ($18\,\mathrm{s} < t \leq 25\,\mathrm{s}$), and \textbf{Expert} ($t \leq 18\,\mathrm{s}$).%\homa{add that we divided the participants into three proficiency levels based on their task completion time of their last trial under normal conditions. Provide the thresholds for N, I, E}

The experiments involved performing the PT task under four primary network conditions: Normal (no degradation), Packet Loss (PLM), Delay (DLM), and Communication Loss (CLM). For each of these degradation types, participants were exposed to three severity levels: low (1), medium (2), high (3). The orders of the degradation types and their severity levels were randomized to eliminate potential biases. Each participant performed 12 trials, including 72 peg transfers. After completing all trials for a given degradation type, participants completed a NASA Task Load Index (TLX) questionnaire \cite{hart2006} to express their perceived workload and experience across varying network conditions. 
% complementing the objective metrics described in Section \ref{Metrics}.

%These subjective scores were used to evaluate user perceived workload and adaptation across varying network conditions, complementing the objective metrics described in Section \ref{Metrics}.

% As clutch usage and motion length has been shown to be indicative of user skill level~\cite{clutch1}, we classified participants into three proficiency groups based on their average clutch usage during PT trials under normal conditions: \textbf{Novice} ($n > 10$), \textbf{Intermediate} ($5 < n \leq 10$), and \textbf{Expert} ($n \leq 5$).  
%\homa{report the total size of data we collected, including number of trials per participants, number of transfers,etc. You had a figure..}

%specifically motion length, clutch usage, and number of errors and failures,
% \subsection{Hardware Configuration}
% The surgeon-side console was connected to a workstation running Windows 7 with an Intel core i7 CPU. The patient-side simulation environment ran on a separate PC with an Ubuntu 20.04 OS, an Intel core i9 CPU, and an NVIDIA GeForce RTX 3090 GPU. The two systems were connected via a local Ethernet network, through which the NetFI emulator intercepted and manipulated the traffic.

\subsection{Network Degradation Parameters and Characteristics}
\label{net_params}
The network conditions for the user study were generated using the stochastic models and optimization process described in Section III. Table \ref{tab:impairment_params_compact} lists the specific parameters optimized (Column 4) for each degradation type and severity level to achieve the target QoS degradation within 2\% error. Under normal conditions, delay is kept within 5ms, and packet loss under 0.01\%, which is considered ideal for a 5G teleoperation infrastructure~\cite{patel2025}. 

\begin{table}[hb!]
\vspace{-0.5em}
\centering
\caption{Key parameters for emulated network degradations}
\label{tab:impairment_params_compact}
\resizebox{\columnwidth}{!}{%
\begin{tabular}{@{}p{2.5cm}lll@{}} % <-- narrower first column
\toprule
\textbf{Degradation Type} & \textbf{Severity} & \textbf{Target} & \textbf{Model Parameters $\theta$} \\ \midrule
\multirow{3}{*}{Packet Loss} & PLM-1 & 10\% Loss & $\alpha_0=4.43, \lambda_0=1.64$; $\alpha_1=4.97, \lambda_1=0.27$ \\
 & PLM-2 & 30\% Loss & $\alpha_0=4.60, \lambda_0=3.55$; $\alpha_1=3.90, \lambda_1=1.47$ \\
 & PLM-3 & 50\% Loss & $\alpha_0=3.86, \lambda_0=3.32$; $\alpha_1=3.80, \lambda_1=6.05$ \\ \midrule
\multirow{3}{*}{Delay} & DLM-1 & 100 ms & $D_{\text{min}}=70, w=[.73,.27], \lambda=[.027,.085]$ \\
 & DLM-2 & 300 ms & $D_{\text{min}}=210, w=[.892,.108], \lambda=[.011,.016]$ \\
 & DLM-3 & 500 ms & $D_{\text{min}}=406, w=[.895,.105], \lambda=[.010,.015]$ \\ \midrule
\multirow{3}{*}{Communication Loss} & CLM-1 & 10\% Loss & $L \sim U(0, 3000)\text{ms}, C=8500\text{ms}$ \\
 & CLM-2 & 30\% Loss & $L \sim U(0, 1000)\text{ms}, C=1000\text{ms}$ \\
 & CLM-3 & 50\% Loss & $L \sim U(0, 2000)\text{ms}, C=1000\text{ms}$ \\ \bottomrule
\end{tabular}%
}
\vspace{-1em}
\end{table}

%\textbf{Choice of Target Degradation Values:} 
The target degradation levels reflect diverse telecommunication conditions and use cases. Low levels represent transient sub-optimal scenarios common in commercial networks, medium levels correspond to conditions that may degrade user performance and patient outcomes~\cite{QoS_Stats1, QoS_Stats2, QoS_Stats3, packetloss_pareto}, and high levels emulate adverse environments such as deep-sea~\cite{Doarn} or battlefield networks~\cite{garcia}. Target communication loss and packet loss rates are selected to be similar (10\%, 30\%, 50\%) to capture how they affect user performance and experience differently. Additionally, parameters of CLM-2 are selected such that it simulates frequent, short and unpredictable losses, while CLM-3 represents longer, more frequent, but more stable outages.

\subsection{Data Collection and Objective Metrics}
\label{Metrics}
During each trial, we recorded comprehensive data streams, including high-frequency kinematic data from the 7-DoF console MTMs and simulated PSMs, clutch pedal state, and a 60Hz video of the simulation environment. We manually annotated all the trials for MPs, errors, and failures by reviewing the recorded videos. This resulted in an annotated multimodal dataset of \textbf{180 trials} (total of \textbf{1,080 peg transfers}), which we used for post-hoc analysis of objective and subjective performance.

\noindent \textbf{Performance Metrics:} We analyzed three primary objective performance metrics to reveal changes in operator behavior under different types of network degradations:
\begin{itemize}
    \item \textbf{Completion Time:} The total time taken to successfully complete the intended action (a transfer or a single MP).
    \item \textbf{Motion Length:} The path length traveled by the console manipulators and the simulated PSM end-effectors. 
    \item \textbf{Clutch Usage:} The number of clutch pedal activations, indicating user adjustments and re-orientations.
\end{itemize}

% \noindent \textbf{Subjective Metrics:} User experience was quantified using the NASA-TLX questionnaire, which measures perceived workload across six dimensions in Fig. \ref{NASA TLX}.

\noindent \textbf{Safety Metrics:}
\label{Modes}
To objectively assess the impact on task safety, we analyzed the occurrence of errors and failures in each trial and across different MPs as defined below:

\begin{itemize}
    \item\textbf{Errors} are events that result in deviation from the ideal task workflow, but are recoverable:
\begin{enumerate}
    \item \textbf{Multiple Attempts:} Repeating a single MP more than twice.
    \item \textbf{Object Drop:} Dropping the peg in a recoverable orientation (i.e., hole facing upwards).
    \item \textbf{Collision:} Crash between the grasper and other objects (peg board, other grasper, etc.).
    \item \textbf{Out of View:} The grasper tip moving completely out of the camera field of view.
\end{enumerate}

\item\textbf{Failures} are irrecoverable events, resulting from errors:
\begin{enumerate}
    \item \textbf{On-Board Drop:} Dropping the peg on its side on the board, making it irrecoverable.
    \item \textbf{Off-Board Drop:} Dropping the peg completely off the peg board.
    \item \textbf{Incorrect Placement:} Placing the peg on a pole of a different color.
\end{enumerate}
\end{itemize}

Based on objective performance and safety metrics under normal trials,  we classified all participants evenly into three proficiency groups: \textbf{Novice}, \textbf{Intermediate}, and \textbf{Expert}.

\section{EXPERIMENTAL RESULTS}
\subsection{Completion Time}

\begin{table*}[!htb]
\centering
\caption{Mean and standard deviation of Completion Time (s) for single peg transfer and individual MPs under network degradations across participants {\footnotesize (bold values indicate changes exceeding 1.5s in MPs and 5s in the overall transfer from normal; * indicates a statistically significant difference between the normal condition and each network degradation based on a paired t-test $(p < 0.001)$)}}
\resizebox{\textwidth}{!}{%
\begin{tabular}{lcccccccccc}
\toprule
\textbf{Conditions} & \textbf{Transfer} & \textbf{MP1} & \textbf{MP2} & \textbf{MP3} & \textbf{MP4} & \textbf{MP5} & \textbf{MP6} & \textbf{MP7} & \textbf{MP8} & \textbf{MP9} \\
\midrule
Normal & 22.1{\scriptsize($\pm$5.6)} & 6.0{\scriptsize($\pm$2.1)} & 1.3{\scriptsize($\pm$0.5)} & 1.6{\scriptsize($\pm$0.4)} & 5.6{\scriptsize($\pm$1.7)} & 1.3{\scriptsize($\pm$0.4)} & 0.9{\scriptsize($\pm$0.3)} & 0.8{\scriptsize($\pm$0.2)} & 3.5{\scriptsize($\pm$1.1)} & 1.2{\scriptsize($\pm$0.4)} \\
\midrule
PLM-1 & 24.4{\scriptsize($\pm$8.3)} & 7.5{\scriptsize($\pm$4.0)} & 1.2{\scriptsize($\pm$0.5)} & 1.9{\scriptsize($\pm$0.9)} & 6.4{\scriptsize($\pm$3.4)} & 1.3{\scriptsize($\pm$0.4)} & 0.9{\scriptsize($\pm$0.3)} & 0.8{\scriptsize($\pm$0.3)} & 3.7{\scriptsize($\pm$1.6)} & 1.1{\scriptsize($\pm$0.3)} \\
PLM-2 & 26.6{\scriptsize($\pm$7.4)} & 6.9{\scriptsize($\pm$2.9)} & 1.3{\scriptsize($\pm$0.7)} & 1.9{\scriptsize($\pm$0.8)} & \textbf{7.9{\scriptsize($\pm$3.2)}} & 1.3{\scriptsize($\pm$0.6)} & 1.0{\scriptsize($\pm$0.4)} & 0.9{\scriptsize($\pm$0.3)} & 4.1{\scriptsize($\pm$2.0)} & 1.1{\scriptsize($\pm$0.3)} \\
PLM-3 & \textbf{36.2$^{*}${\scriptsize($\pm$10.7)}} & \textbf{10.7$^{*}${\scriptsize($\pm$5.0)}} & 1.3{\scriptsize($\pm$0.5)} & 2.8{\scriptsize($\pm$1.4)} & \textbf{9.8$^{*}${\scriptsize($\pm$3.7)}} & 1.6{\scriptsize($\pm$0.6)} & 1.3{\scriptsize($\pm$0.6)} & 1.2{\scriptsize($\pm$0.4)} & \textbf{5.6$^{*}${\scriptsize($\pm$1.6)}} & 1.1{\scriptsize($\pm$0.2)} \\
\midrule
DLM-1 & \textbf{27.1{\scriptsize($\pm$9.7)}} & 7.0{\scriptsize($\pm$3.1)} & 1.2{\scriptsize($\pm$0.4)} & 1.8{\scriptsize($\pm$0.8)} & \textbf{7.4{\scriptsize($\pm$4.0)}} & 1.4{\scriptsize($\pm$0.6)} & 1.0{\scriptsize($\pm$0.4)} & 1.0{\scriptsize($\pm$0.3)} & 4.5{\scriptsize($\pm$1.8)} & 1.3{\scriptsize($\pm$0.5)} \\
DLM-2 & \textbf{34.1$^{*}${\scriptsize($\pm$6.5)}} & \textbf{10.4$^{*}${\scriptsize($\pm$3.9)}} & 1.6{\scriptsize($\pm$1.1)} & 2.5{\scriptsize($\pm$0.9)} & \textbf{8.8$^{*}${\scriptsize($\pm$2.9)}} & 1.6{\scriptsize($\pm$0.6)} & 1.2{\scriptsize($\pm$0.4)} & 1.4{\scriptsize($\pm$0.5)} & \textbf{6.1$^{*}${\scriptsize($\pm$1.9)}} & 1.6{\scriptsize($\pm$0.6)} \\
DLM-3 & \textbf{47.9$^{*}${\scriptsize($\pm$12.9)}} & \textbf{14.4$^{*}${\scriptsize($\pm$5.4)}} & 2.1{\scriptsize($\pm$1.2)} & 2.9{\scriptsize($\pm$1.4)} & \textbf{12.9$^{*}${\scriptsize($\pm$3.9)}} & 2.0{\scriptsize($\pm$1.2)} & 1.6{\scriptsize($\pm$0.4)} & 1.9{\scriptsize($\pm$0.7)} & \textbf{9.1$^{*}${\scriptsize($\pm$2.9)}} & 1.7{\scriptsize($\pm$0.6)} \\
\midrule
CLM-1 & \textbf{28.9$^{*}${\scriptsize($\pm$8.9)}} & \textbf{8.2{\scriptsize($\pm$3.7)}} & 1.4{\scriptsize($\pm$0.7)} & 2.3{\scriptsize($\pm$1.3)} & \textbf{7.6{\scriptsize($\pm$3.4)}} & 1.4{\scriptsize($\pm$0.8)} & 1.2{\scriptsize($\pm$0.4)} & 1.2{\scriptsize($\pm$0.6)} & 4.7{\scriptsize($\pm$1.9)} & 1.2{\scriptsize($\pm$0.6)} \\
CLM-2 & \textbf{47.4$^{*}${\scriptsize($\pm$11.0)}} & \textbf{13.5$^{*}${\scriptsize($\pm$4.8)}} & 1.9{\scriptsize($\pm$0.9)} & 3.9{\scriptsize($\pm$2.2)} & \textbf{14.5$^{*}${\scriptsize($\pm$4.9)}} & 1.6{\scriptsize($\pm$0.6)} & 1.4{\scriptsize($\pm$0.7)} & 1.5{\scriptsize($\pm$0.5)} & \textbf{7.3$^{*}${\scriptsize($\pm$1.8)}} & 1.5{\scriptsize($\pm$0.5)} \\
CLM-3 & \textbf{64.3$^{*}${\scriptsize($\pm$25.4)}} & \textbf{17.4$^{*}${\scriptsize($\pm$8.4)}} & 2.7{\scriptsize($\pm$2.7)} & \textbf{4.7$^{*}${\scriptsize($\pm$2.6)}} & \textbf{20.2$^{*}${\scriptsize($\pm$10.4)}} & \textbf{2.9{\scriptsize($\pm$2.1)}} & 2.2{\scriptsize($\pm$0.9)} & 2.3{\scriptsize($\pm$0.9)} & \textbf{10.2$^{*}${\scriptsize($\pm$3.1)}} & 2.0{\scriptsize($\pm$1.1)} \\
\bottomrule
\end{tabular}
}
\label{tab:CompletionTime}
\vspace{-1.5em}
\end{table*}
As shown in Table \ref{tab:CompletionTime}, \textbf{all Touch MPs (MP1, MP4, MP8), are most affected by three intensified network degradations}, demonstrating substantial value changes that contribute significantly to the overall increase in total transfer time. Notably, MP4 is impacted even under low-level delay (DLM-1) and 30\% packet loss (PLM-2), making it the most challenging MP due to its involvement in a coordinated two-handed position and orientation adjustment.%, which further confirms findings from previous studies \cite{Hutchinson_2022, Hutchinson_2023}. 

In contrast to packet loss, communication loss introduces prolonged periods of packet drop that can completely stall robotic motion. Consequently, it leads to longer Touch completion times across all severity levels, and \textbf{even smaller actions (MP3 and MP5) are affected under the highest level of communication loss (CLM-3)}. Delay, on the other hand, increases the transmission time of each packet from the surgeon side to the patient side. To maintain accurate and precise robotic control under delayed feedback, users are forced to slow down %their movements. 
As a result, completion time of overall transfer and larger motion segments increase substantially.  

% Another key observation is that all network QoS degradations increase inconsistency in user performance, as reflected by a significant rise in the standard deviation of completion time with worsening severity levels. \textbf{This indicates that network degradation not only deteriorates overall performance but also amplifies the variability between users}. 

\subsection{Motion Length}

\begin{figure}
    \centering
    \includegraphics[width=1 \linewidth]{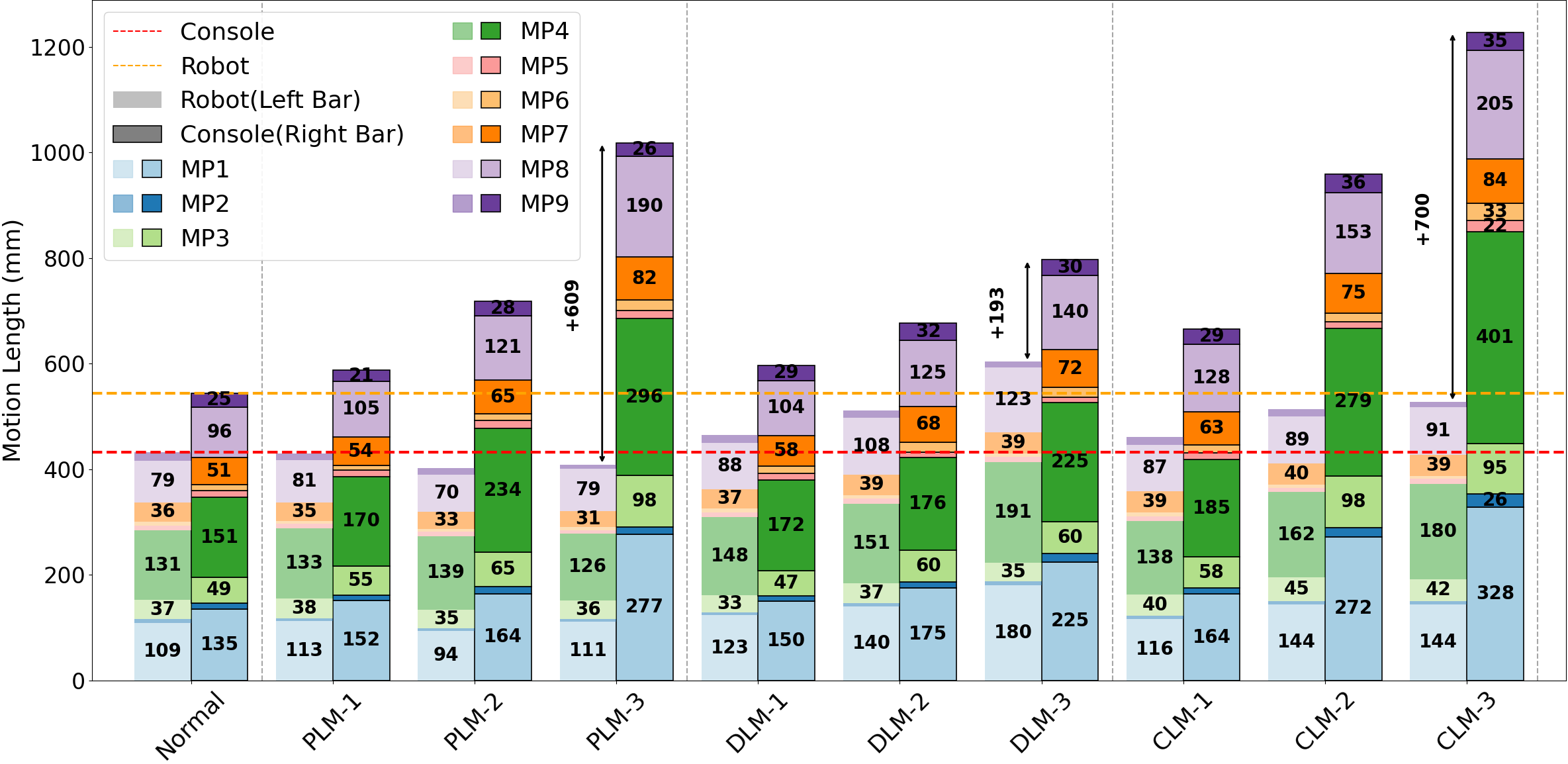}
    \caption{Comparison of the average motion length of the PSMs and MTMs at the MP level for all participants under varying levels of packet loss, delay, and communication loss.}
    \label{Motion Length Stacked Bar Chart} 
    \vspace{-1.5em}
\end{figure}

As shown in Fig. \ref{Motion Length Stacked Bar Chart}, increasing severity levels of all network degradations result in a clear upward trend in the overall average transfer motion length of the MTMs. In contrast, the total motion length in robot space shows no significant variation across packet loss levels relative to the normal condition ($p > 0.05$). While overall robot motion increases under higher delay and communication loss, smaller MPs contribute minimally.

Moreover, with higher packet and communication loss rates, the robot and console space motion lengths of overall transfer diverge further because lost commands force users to perform additional compensatory movements. \textbf{Although loss rates are similar for packet and communication loss (10\%, 30\%, 50\%), hand movements are larger under communication loss, due to more oscillatory adjustments by the user}. This is because communication loss not only interrupts intended actions, but also leads the participants to make small movements to determine whether the communication has been restored. This effect is particularly evident in larger MPs, such as MP1, MP4, and MP8. 

In contrast, delay preserves all packets but can distort the operator’s intended control trajectory. Excessive movement speed under high-delay conditions may cause overshooting, which accounts for the increased motion length observed in both the robot and console spaces. Also, console manipulator movements at DLM-3 is shorter than those observed under the most severe packet loss and communication loss conditions.

\subsection{Clutch Usage}

Table~\ref{tab:PedalUsage} shows that both packet and communication loss lead to increased clutch usage across all participant groups. However, \textbf{novices are the most affected by higher levels of communication loss, as indicated by highest increase in clutch usage for adjustment and re-orientation}. On the other hand,  experts demonstrate proficiency in managing uncertain and extended communication outages through more stable clutch usage. Additionally, MPs involving larger movements (MP1 and MP4) require more frequent clutch use under packet loss and communication loss. However, smaller MPs show no significant change because their limited movements can be completed within the surgeon console’s existing workspace without re-adjustment.

\textbf{Interestingly, delay does not significantly increase pedal usage across all three groups $(p > 0.05)$}. This is attributed to participants instinctively slowing their movements to maintain control, and because no command data are lost, the need for spatial adjustment via clutch remains largely unchanged.

%both packet loss and communication loss significantly increase pedal usage, \textbf{particularly for MPs involving larger movements like \textit{MP1} and \textit{MP4}}. This increase is most pronounced under medium (CLM-2) and high (CLM-3) communication loss, where the uncertain and probabilistic nature of the outages leads to unpredictable instrument movements that counter the user's intended actions, necessitating frequent re-clutching. Severe packet loss (PLM-3) also leads to a significant rise in pedal presses as users must often compensate for lost commands to realign their intended movements. In contrast, \textbf{for smaller, more intricate MPs, no significant change in pedal usage is observed}. This is likely because the required movements are small enough to be performed within the surgeon console's existing workspace without re-adjustment. 

\begin{table}[!htb]
\centering
\caption{Mean and standard deviation of clutch usage for the six-peg transfer of all participants and each proficiency group {\footnotesize(bold values indicate the highest value within each group and arrows denote distinct trends relative to the preceding severity level)}}

\resizebox{\columnwidth}{!}{%
\begin{tabular}{lcccc}
\toprule
\textbf{Conditions} & \textbf{All} & \textbf{Novice} & \textbf{Intermediate} & \textbf{Expert}\\
\midrule
Normal & 9 {\scriptsize($\pm$4.4)} & 14 {\scriptsize($\pm$1.7)}& 8 {\scriptsize($\pm$0.8)}& 4 {\scriptsize($\pm$1.0)} \\
\midrule
PLM-1 & 9 {\scriptsize($\pm$6.9)}&
13 {\scriptsize($\pm$9.7)}& 10 {\scriptsize($\pm$3.3)}& 4 {\scriptsize($\pm$1.7)}\\
PLM-2 & 13 {\scriptsize($\pm$6.6)} & 16 {\scriptsize($\pm$5.0)} & 16 {\scriptsize($\pm$6.3)} & 6 {\scriptsize($\pm$2.3)}\\
PLM-3 & \textbf{23} {\scriptsize($\pm$14.2)} & \textbf{36} {\scriptsize($\pm$16.9)} & \textbf{22} {\scriptsize($\pm$5.3)} & \textbf{12} {\scriptsize($\pm$1.3)}\\
\midrule
DLM-1 & 8 {\scriptsize($\pm$5.3)} & 14 {\scriptsize($\pm$3.7)} & 8 {\scriptsize($\pm$3.7)} & 3 {\scriptsize($\pm$1.0)}\\
DLM-2 & 10 {\scriptsize($\pm$5.3)} & 15 {\scriptsize($\pm$6.1)} & 9 {\scriptsize($\pm$2.1)} & 6 {\scriptsize($\pm$2.2)} \\
DLM-3 & \textbf{11} {\scriptsize($\pm$5.6)} & \textbf{17} {\scriptsize($\pm$4.6)} & \textbf{9} {\scriptsize($\pm$2.3)} & \textbf{6} {\scriptsize($\pm$2.0)} \\
\midrule
CLM-1 & 10 {\scriptsize($\pm$5.9)} & 17 {\scriptsize($\pm$4.4)} & 9 {\scriptsize($\pm$3.9)} & 5 {\scriptsize($\pm$1.3)} \\
CLM-2 & 20 {\scriptsize($\pm$9.3)} & 29 {\scriptsize($\pm$10.0)~$\uparrow$} & 17 {\scriptsize($\pm$4.5)} & \textbf{15} {\scriptsize($\pm$5.4)~$\uparrow$} \\
CLM-3 & \textbf{25} {\scriptsize($\pm$17.0)} & \textbf{44} {\scriptsize($\pm$15.6)~$\uparrow$} & \textbf{18} {\scriptsize($\pm$5.2)} & 12 {\scriptsize($\pm$2.9)~$\downarrow$}\\ 
\bottomrule
\end{tabular}
}
\label{tab:PedalUsage}
\vspace{-1.5em}
\end{table}

\subsection{Errors and Failures}
Table~\ref{tab:ErrorFailure} shows that degraded network QoS generally leads to more errors and a lower task success rates. The success rate deteriorates under all network conditions, with severe delay (DLM-3) causing the largest drop (around 9\%). Communication loss at a 30\% average rate (CLM-2) also has a substantial impact; its frequent but brief loss windows make it highly unpredictable, making it difficult for users to plan movements effectively and leading to more mistakes (see subsection \ref{net_params}). Most failures resulted from dropping the peg on the board in an unrecoverable position (Failure Mode 1), with only four instances of placing the peg on the wrong pole (Failure Mode 3).

Analyzing specific error types provides deeper insights into \textit{RQ2}. The number of \textit{multiple attempts} errors correlates with increasing levels of packet loss and delay. The unpredictable nature of CLM-2 also contributes to a large share of these errors. \textbf{These errors are most common in MPs involving touching and grasping (MP1, MP4, MP5), indicating that network degradations make precise manipulation more challenging}. \textbf{\textit{Object drops} primarily occur during MP3 (Untouching the peg) and MP4 (Touching for exchange), which often involve faster and preciser movements}. Consequently, lower and medium levels of network degradations are more likely to cause drops, as users tend to move more cautiously under severe conditions. \textbf{Around 70\% \textit{collisions} occur during MP4, when the instruments are closest. Jerky, overshoot-prone movements under high delay and unpredictable motions during communication loss are the primary causes}. Finally, an \textit{out of view} error occurred only under delay, caused by a large overshoot during an untouch MP.

Errors and failures were further analyzed by proficiency group. With the exception of a single intermediate participant who experienced three drop failures under PLM-1, \textbf{novices were significantly more prone to errors than experts and intermediates} as network degradation severity increased, particularly in \textit{multiple attempts} and \textit{collision} errors. 
%The error and failure mode numbers correspond to those defined in Section \ref{Modes}.

\begin{table}[!t]
    \centering
    \caption{Success Rate, Error Counts, and Failure Counts. \textbf{Error Modes}: (1) Multiple Attempts; (2) Object Drop; (3) Collision; (4) Out of View; \textbf{Failure Modes}: (1) On-Board Drop; (2) Off-Board Drop; (3) Incorrect Placement {\footnotesize (rows for the most error-prone network degradation levels are in bold)}}
    \label{tab:ErrorFailure}
    \resizebox{\columnwidth}{!}{
    \begin{tabular}{lccccccccc}
        \toprule
        Conditions & Success Rate & \multicolumn{4}{c}{Error Counts} & \multicolumn{3}{c}{Failure Counts} \\
        & (\%) & (1) & (2) & (3) & (4) & (1) & (2) & (3) \\
        \midrule
        Normal-1 & 100.0 (90/90) & 8 & 3 & 0 & 0 & 0 & 0 & 0 \\
        Normal-2 & 100.0 (90/90) & 9 & 6 & 3 & 0 & 0 & 0 & 0 \\
        Normal-3 & 98.8 (86/87) & 4 & 6 & 2 & 0 & 1 & 0 & 0 \\
        \midrule
        PLM-1 & 93.2 (82/88) & 11 & 12 & 5 & 0 & 6 & 0 & 0 \\
        PLM-2 & 97.8 (88/90) & 15 & 5 & 7 & 0 & 1 & 1 & 0 \\
        \textbf{PLM-3} & 95.5 (85/89) & \textbf{17} & \textbf{11} & \textbf{6} & 0 & \textbf{4} & 0 & 0 \\
        \midrule
        DLM-1 & 95.6 (86/90) & 6 & 8 & 6 & 0 & 3 & 1 & 0 \\
        DLM-2 & 92.2 (83/90) & 13 & 18 & 14 & 2 & 4 & 1 & 2 \\
        \textbf{DLM-3} & 92.2 (83/90) & \textbf{37} & \textbf{12} & \textbf{26} & 0 & \textbf{5} & 0 & \textbf{2} \\
        \midrule
        CLM-1 & 100.0 (90/90) & 5 & 9 & 3 & 0 & 0 & 0 & 0 \\
        \textbf{CLM-2} & 95.5 (85/89) & \textbf{26} & \textbf{21} & \textbf{20} & 0 & \textbf{4} & 0 & 0 \\
        CLM-3 & 97.7 (87/89) & 13 & 11 & 13 & 0 & 2 & 0 & 0 \\
        \bottomrule
        \multicolumn{2}{c}{Most Susceptible MPs} & MP1 & MP3 \& MP4 & MP4 & MP7 & MP4 & MP8 & MP4
    \end{tabular}
    }
    \vspace{-2em}
\end{table}

\subsection{Perceived Task Load}
\begin{figure}
    \centering
    \includegraphics[width=0.87\linewidth]{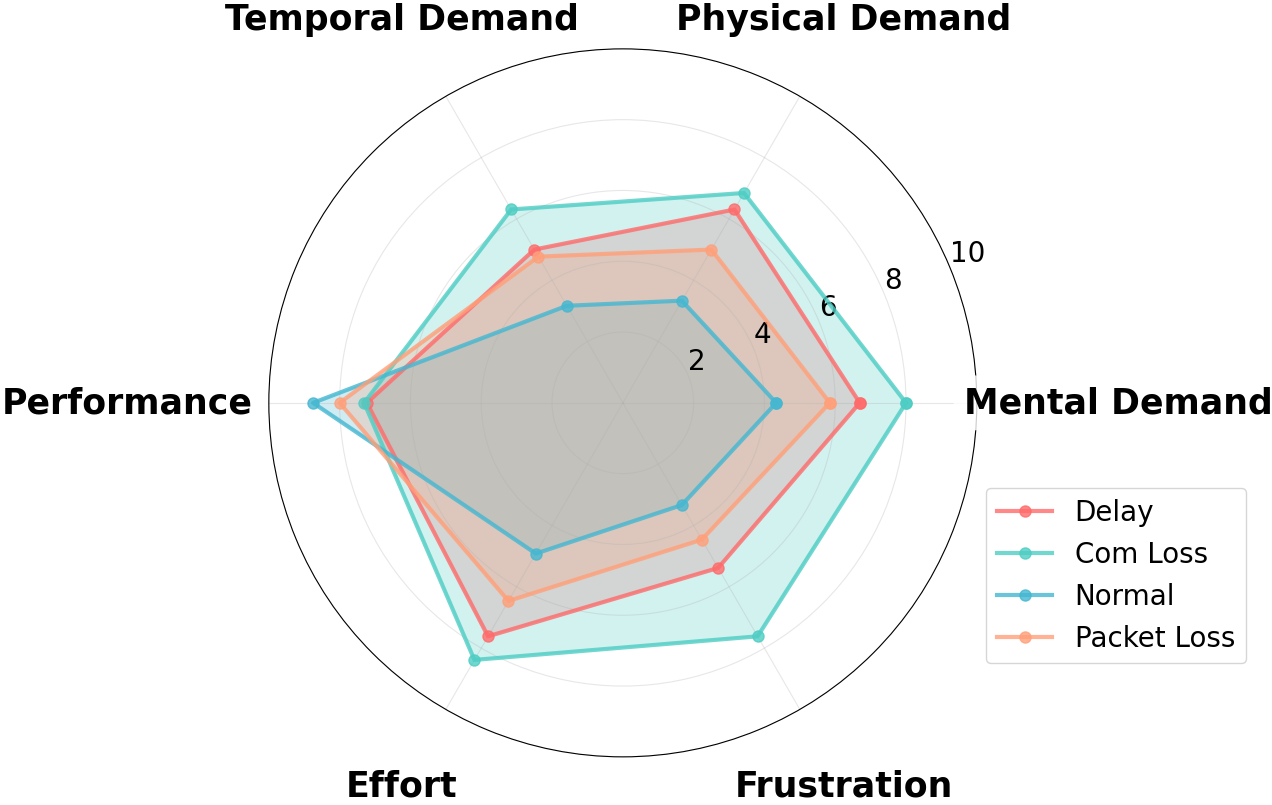}
    \caption{TLX Perceived Task Load across all participants}
    \label{NASA TLX}
    \vspace{-1.7em}
\end{figure}

Users rated our teleoperation system with an aggregate SUS score of 87.5/100, indicating excellent performance and helping to isolate the effects of network degradations from system shortcomings. The NASA-TLX results (Fig. ~\ref{NASA TLX}) reveal how degradations affect subjective experience (\textit{RQ3}). We observed a clear correlation between objective performance and perceived workload measurements. \textit{Delay} and \textit{Communication Loss}, which increased completion times, were also linked to higher \textit{Frustration}, \textit{Effort}, and \textit{Physical Demand}.

% Communication loss was especially taxing, producing the highest \textit{Mental Demand} due to the need for prediction and constant connection monitoring. It was also rated more frustrating than other conditions. The general increase in \textit{Effort} and \textit{Physical Demand} across all degradations stemmed from compensatory actions. \textbf{Under packet and communication loss, users repeated motions and used the clutch pedal more frequently, reflected in both higher physical demand and increased pedal usage}.

% Self-assessed \textit{Performance} scores were lowest under delay and communication loss, aligning with higher error rates and showing strong correlation between errors and perceived accomplishment. Finally, across proficiency groups, \textbf{the most significant disparity appeared in \textit{Physical and Temporal Demand} between experts and novices}, with less proficient users finding degradations more demanding due to longer completion times and larger hand movements.%\textbf{Recovering from these errors further increases the physical and mental demands on the user, completing the feedback loop between poor network quality, degraded performance, and heightened perceived workload}.

To investigate the relationship between objective performance and subjective workload, we performed a Pearson correlation analysis between kinematic metrics and NASA-TLX dimensions. Statistical analysis revealed that completion time is significantly correlated with Physical Demand ($r=0.38, p<0.01$), Frustration ($r=0.37, p<0.01$), and Mental Demand ($r=0.36, p<0.01$), indicating that longer task durations under network degradation directly increase perceived strain. Subjective Performance scores were negatively correlated with both Error Count ($r=-0.40, p<0.01$) and Motion Length ($r=-0.38, p<0.01$), showing that operators were acutely aware of reduced efficiency and precision. Furthermore, Physical Demand positively correlated with Clutch Usage ($r=0.34, p<0.01$), reflecting the increased effort required for frequent spatial re-adjustments.

\section{CONCLUSIONS}
This paper presented a systematic investigation into the effects of network QoS degradations on telesurgery through a comprehensive user study using a novel, model-based fault injection framework and an open-source simulation platform. Our findings demonstrate that not all network degradations are equal. Communication loss and severe delay disproportionately degrade task performance, increase error rates, and heighten operator workload compared to even significant levels of packet loss. Decomposing surgical tasks into MPs allowed us to identify specific sub-tasks that are acutely sensitive to network instability. These granular insights could help design of context-aware autonomy to assist operators during vulnerable segments of a procedure. 

However, our study is primarily limited by its use of non-expert participants and a simulated, rigid-body task that lacks haptic feedback and the complexities of soft-tissue manipulation. Future work will address these limitations by validating our findings with experienced surgeons in more realistic and complex scenarios. We also plan to incorporate dynamic network factors, including jitter, bandwidth variability, and video distortions, to more accurately model real-world telesurgical environments. The telesurgical simulation framework, tools, and dataset resulted from this work provide a critical foundation for advancing research on the reliability, safety, and autonomy of telesurgery, particularly rigorous development and testing of intelligent mitigation strategies.%, such as predictive filtering, shared-autonomy frameworks, and adaptive user interfaces, that will be essential for ensuring the safety and efficacy of the next generation of telesurgical systems.

%                                   % the textheight of the last page by a suitable amount.
%                                   % This command does not take effect until the next page
%                                   % so it should come on the page before the last. Make
%                                   % sure that you do not shorten the textheight too much.

% \section*{ACKNOWLEDGMENT}

%\bibliographystyle{ieeetr}
\bibliographystyle{IEEEtran}
\bibliography{ref}

\end{document}